\documentclass[runningheads]{llncs}
\usepackage{graphicx}
\usepackage{mathtools}
\usepackage{cite}
\usepackage{amsmath,amssymb,amsfonts}
\usepackage{algorithmic}
\usepackage{graphicx}
\usepackage{textcomp}
\usepackage{xcolor}
\usepackage{multirow}
\usepackage{todonotes}
\def\BibTeX{{\rm B\kern-.05em{\sc i\kern-.025em b}\kern-.08em
    T\kern-.1667em\lower.7ex\hbox{E}\kern-.125emX}}

\newcommand{\eqDM}{\emph{MMM}}
\newcommand{\Multifair}{MFBPP}
\newcommand{\mf}{\emph{multi-discrimination}}
\newcommand{\sd}{\emph{mono-discrimination}}
\newcommand{\multimax}{\emph{Multi-Max Mistreatment}}

\begin{document}
\title{Multi-fairness under
class-imbalance}
%
%
\author{Arjun Roy\inst{1,2}\orcidID{0000-0002-4279-9442} \and
Vasileios Iosifidis\inst{2}\orcidID{0000-0002-3005-450} \and
Eirini Ntoutsi\inst{1}\orcidID{0000-0002-3005-4507}}
\authorrunning{A. Roy et al.}
%
\institute{Institute of Computer Science, Free University of Berlin, Germany \email{\{arjun.roy,eirini.ntoutsi\}@fu-berlin.de}\and
L3S Research Center, Leibniz University Hannover, Germany
\\\email{iosifidis@l3s.de} 
}
\maketitle              
\begin{abstract}
Recent studies showed that datasets used in fairness-aware machine learning for multiple protected attributes (referred to as \mf~hereafter) are often imbalanced. The \emph{class-imbalance} problem is more severe for the protected group in the critical minority class
(e.g., \emph{female +, non-white +, etc.}). Still, existing methods focus only on the overall error-discrimination trade-off, ignoring the imbalance problem, and thus they amplify the prevalent bias in the minority classes. 
To solve the combined problem of \mf{} and class-imbalance we introduce a new fairness measure, \multimax{} (\eqDM), which considers both (multi-attribute) protected group and class membership of instances to measure discrimination. To solve the combined problem, we propose  Multi-Fair Boosting Post Pareto (\Multifair)~a boosting approach that incorporates \eqDM-costs in the distribution update and post-training, 
selects the optimal trade-off among accurate, class-balanced, and \emph{fair} solutions. 
The experimental results show the superiority of our approach against state-of-the-art methods in producing the best balanced performance across groups and classes and the best accuracy for the protected groups in the minority class. 
\keywords{Multi-discrimination  \and Class-imbalance \and Boosting.}
\end{abstract}

\section{Introduction}
\label{sec:intro}
There are growing concerns about the potential discrimination and unfairness of Machine Learning (ML) models in areas of high societal impact like recidivism, job hiring and loan credit. 
Over the last years a growing body of  works has been proposed 
to address the problem of fairness and algorithmic discrimination~\cite{NtoutsiEtAl:WIDM:2020}. The vast majority of  fairness-aware ML approaches however, assumes that discrimination is due to a single protected attribute e.g., only race or only gender (referred hereafter as \emph{mono-discrimination}).
In reality though, the roots of discrimination can be ascribed to \emph{multiple protected attributes} (referred hereafter as \mf\footnote{Through the paper we use the terms ``multi-discrimination'' and ``multi-fairness'' interchangeably.}), e.g., a combination of race, gender and age~\cite{makkonen2002multiple}. 

The problem of multi-discrimination has attracted attention recently and several approaches to multi-fairness have been proposed~\cite{agarwal2018reductions,zafar2019fairness,kang2021multifair_MI,martinezminimax,yang2020fairness_overlap}. 
\color{red}
\color{black}
 However, none of the existing \mf~ methods considers class-imbalance and the problem arising out of it.
Studies~\cite{hu2020fairnn,IosNto:CIKM:19,le2022oursurvey} showed that many datasets used in fairness-aware ML research are \emph{class-imbalanced}, i.e., they contain a disproportionately larger amount of instances from the majority class (typically called negative ``-'' class) comparing to the minority class (typically called positive ``+'' class).
The imbalance is even more pronounced in protected groups like female (vis-a-vis male), non-white (vis-a-vis white) etc.   
Table~\ref{tbl:data} highlights the problem in three real-world datasets, which are widely used to evaluate fair-ML algorithms~\cite{le2022oursurvey}. 
\begin{table*}
   \caption{Overview of class imbalance ratio (CIR) and protected:non-protected group imbalance ratio in the minority ``+" class (GIR) for different protected attributes.
   }\label{tab:data}
   \vspace{4mm}
      \resizebox{\columnwidth}{!}{%
    \begin{tabular}{|c|c|c|c|c|}
    \hline
       \textbf{Data}  & $n$  & \textbf{Minority (+) class} & \textbf{CIR (+:-)} & \textbf{GIR (Prot. : Non-prot) in ``+" class}\\
         \hline
         Adult & 45K  & $>50k$ & ~1:3 &  \begin{tabular}[c]{@{}c@{}} Race: (1:6),Sex: (1:2)\end{tabular}\\

         Bank  & 40K & \textit{subscription} & ~1:8.9& \begin{tabular}[c]{@{}c@{}} Marital (1:3), Age (1:23)\end{tabular}\\

         Credit & 30K & \begin{tabular}[c]{@{}c@{}} \textit{default pay.} \end{tabular}&~1.4:3 & \begin{tabular}[c]{@{}c@{}} Sex (1:1.5), Age (1:6), Marital (1:1.5) \end{tabular}\\
         \hline
    \end{tabular}}
    \label{tbl:data}
    \end{table*}
The Class Imbalance Ratio (\textit{CIR}) is the (+/-) ratio in the whole dataset.
For the minority `+' class, we also show the Group Imbalance Ratio (\textit{GIR}) which is the ratio between the protected and non-protected groups, for different protected attributes. 
As seen in Table~\ref{tbl:data}, within the minority `+' class there exist extreme imbalance between the protected and non-protected group. Thus,  within the entire data these protected groups have very less `+' examples. The degree of imbalance varies from attribute to attribute. Thus, giving an uniform and equal importance to tackle discrimination for all the protected attributes may not be sufficient.
In these circumstances, a classifier can be highly accurate even by completely ignoring these protected `+' examples. On the other hand, a fair classifier whose working principle is to minimize the difference between performance of the two groups, can have high error rate on both protected and non-protected `+' examples (i.e., predicting them as `-'). Such a situation may result in an acceptable drop in accuracy (in lieu of fairness), however, may lead to heavy under-performance in the positive outcome of some protected groups.

State-of-the-art \mf~methods~\cite{agarwal2018reductions,martinezminimax,morina2019auditing,kang2021multifair_MI,yang2020fairness_overlap,zafar2019fairness} focus only on error-discrimination trade-off, but 
ignore this precise imbalance problem.  
Also, the evaluation strategy presently used ignores to report on this issue of performance of the worst performing group in the minority class. 
Thus, we need a holistic algorithm approach along with a  
thorough evaluation mechanism to  measure that analyses the 
performance based on overall error, \mf, imbalance, and protected groups in the minority class. In this work, we target the combined combined problem of \mf{} and class-imbalance. Our main contributions are as follows:  

\noindent \textbf{i)} We extend the definition of 
multi-group~\cite{yang2020fairness_overlap} fairness to introduce the notion of \multimax~(\eqDM)\footnote{The term `\emph{multi}' here refers to both multiple attributes and multiple classes.} 
that evaluates discrimination for multiple protected attributes and across different classes. 

\noindent\textbf{ii)} We formulate the \mf{} under class-imbalance problem 
as a multi-faceted problem of finding a \eqDM-fair classifier that achieves low overall error, and minimizes performance differences across the classes and groups, to overcome the problem of underrepresented protected groups in the minority (+) class. 

\noindent\textbf{iii)} We propose Multi-Fair Boosting Post Pareto (\Multifair) algorithm, an in-processing boosting-based approach coupled with a post-processing Pareto Front selection to solve the multiple problems in-hand. 

\noindent\textbf{iv)} We demonstrate an all round evaluation based on accuracy, imbalance, \mf, and accuracy of protected groups in the positive class to show the superiority of our \Multifair{} against various state-of-the art approaches w.r.t. \mf{} under class-imbalance.

\noindent\textbf{v)} We offer a flexible alternative of our model to provide solutions per user needs based on user preferences. 

The rest of the paper is organized as follows: Related work is summarized in Section~\ref{sec:related}. In Section~\ref{sec:ourMeasure} we introduce basic notation and our \multimax ~(\eqDM) fairness measure. Our boosting-based method towards an 
\eqDM-fair classifier is presented in Section~\ref{sec:MMM} and the experimental evaluation in Section~\ref{sec:experiments}. We conclude this work in Section~\ref{sec:conclusions} where we also point to open directions.

\section{Related Work}
\label{sec:related}
In the following, we summarize related work referring to \mf, and imbalanced learning. Notions built around intersectional discrimination~\cite{19icdeIntersectional,kearns2018gerrymandering} is the most common practice to measure \mf. However, such measures suffer from the drawback of clarity in subgroup definition~\cite{fredman2016intersectional} and scarcity in subgroup distribution~\cite{kearns2018gerrymandering}. Recently, works~\cite{kang2021multifair_MI,yang2020fairness_overlap} towards the more operational \mf~measure concerning disjoint groups defined by multiple protected attributes came into light. However, they do not take into account ground truth or class membership which is important to consider in presence of class-imbalance. Our introduced \eqDM~notion, overcomes the issue by considering both class and multi-group membership of the instances to measure \mf.

A few existing approaches~\cite{agarwal2018reductions,martinezminimax,morina2019auditing,kang2021multifair_MI,yang2020fairness_overlap,zafar2019fairness} in supervised learning can handle \mf. 
\cite{zafar2019fairness} introduces fairness-related convex-concave constraints to a logistic regression classifier (FairCons). \cite{agarwal2018reductions} imposes a set of linear fairness constraints on an exponentiated-gradient reduction method (FairLearn). 
\cite{kang2021multifair_MI} tackles the fairness-accuracy trade-off by minimizing mutual information between the learning error and the vectorized multiple protected attributes (MI-Fair). \cite{yang2020fairness_overlap} applies a Bayes-optimal group-fair classifier (W-ERM) to identify the most-dicriminated group. 
Fairness-aware learning as a \emph{mini-max} theory has been already used in the literature \cite{martinezminimax},  
searching for a Pareto efficient solution of a multi-objective problem (MiniMax). 
Recently it has been shown that skewed class distributions can affect the discriminatory behaviour of a model~\cite{hu2020fairnn,fae,IosNto:CIKM:19} in the \sd~set-up. None of the existing \mf~ methods considers class-imbalance. 
Boosting-based approaches have shown their effectiveness in tackling class-imbalance~\cite{chawla2003smoteboost,sun2007cost},  fairness~\cite{hickey2020shapECML20fairness,IosNto:CIKM:19}, and multi-class~\cite{2021multiclass_boost} problems. 
\cite{IosNto:CIKM:19} tackles both fairness and class-imbalance but for a single protected attribute (AdaFair). 

Our proposed \Multifair{} considers both \mf~and class-imbalance to overcome the limitation of multiple underrepresented groups while delivering accurate solutions across the classes.
\section{Basics and \multimax (\eqDM) fairness}
\label{sec:ourMeasure}
We assume a dataset $D=(u^{(i)},s^{(i)},y^{(i)})\sim P$ of $n$ instances drawn from the i.i.d  distribution $P$ over the domain $U\times S \times Y$, where $U$ is the subspace of \emph{non-protected attributes}, $S$ is the subspace of \emph{protected attributes}, and $Y$ is the class attribute. For simplicity, we assume a binary problem: $Y\in \{+,-\} $ 
with `$+$' being the minority ($+$) 
class~\cite{types_of_minority}. 
$U$ and $S$ together define the feature space $X=U\times S$, so $x^{(i)}=(u^{(i)},s^{(i)})$.

Let the protected subspace consist of $k$ protected attributes: $\{S_1, S_2, \cdots, S_k\}$. Each protected attribute is considered to be binary: $\forall_{j=1,\cdots,k} S_j\in \{g_j,\overline{g_j}\}$ and  where $g_j$ and $\overline{g_j}$ represent the \emph{protected group} and the \emph{non-protected group}, respectively w.r.t. protected attribute $S_j$. 
Each group $g_j$ ($\overline{g_j}$) w.r.t. a protected attribute $S_j$ can be further subdivided based on class information into:
\emph{protected positive} $g_{j+}$, 
\emph{protected negative} $g_{j-}$, \emph{non-protected positive} $\overline{g_{j+}}$ and  \emph{non-protected negative} $\overline{g_{j-}}$.

To measure mistreatment in \sd~cases, \cite{zafar2019fairness}~introduced the notion of \emph{Disparate Mistreatment} for a protected feature $j$ as: \begin{equation}
    DM_j=|\delta FNR_j|+|\delta FPR_j|
    \label{eq:DMj}
\end{equation}

where $\delta FNR_j$ ($\delta FPR_j$) is the discrimination w.r.t. $S_j$ in the positive `+' class (respectively, negative `-' class) defined as:
$$\delta FNR_j = ER(g_{j+})-ER(\overline{g_{j+}})$$ 
$$\delta FPR_j =ER(g_{j-})-ER(\overline{g_{j-}})$$

\subsection{\multimax (\eqDM) measure}
\label{sec:MMM}

The Disparate Mistreatment measure (c.f., Equation~\ref{eq:DMj}) fails to focus on per-class discrimination due to the summation operation. 
To ensure fair treatment \emph{across all classes}, for a protected attribute $S_j$, we measure mistreatment as $\max(|\delta FNR_j|,|\delta FPR_j|)$ where the `$\max$' operator  enforces  focus on each of the classes. Moreover,
we want to ensure fair treatment  \emph{across all protected attributes} $S=\{S_1,\cdots,S_k\}$. Our goal is therefore, to focus on the most discriminated group defined based on a protected attribute and a class. 
To this end, we introduce a new \mf~notion, called \multimax ~(\eqDM), that measures the \emph{maximum} discrimination among the protected attributes and for the different classes. 
\begin{definition}
\label{def:eqDM}

The \multimax$(\eqDM_S)$ due to multiple-protected attributes \small{$S=\{S_1, \cdots, S_k\}$}\normalsize~across all classes $Y=\{+,-\}$, is defined as: 
\small
\begin{equation}\label{eq: eqdm}
\begin{multlined}
    \eqDM_{S}= \max_{S_j \in S}\big(\max(|\delta FNR_j|,|\delta FPR_j|)\big)
\end{multlined}
\end{equation}
\normalsize
\end{definition}
where \small$\delta {FNR}_j$\normalsize~and \small{$\delta FPR_j$}\normalsize~measure the mistreatment due to $S_j$ in the (+) and (-) class, respectively. 



\begin{definition}
\label{def:MMM-fairlearner}
Given a tolerance threshold $\mu$, a classifier $f(\cdot)$ is \eqDM-fair \textit{iff}
the maximum mistreatment w.r.t all the protected attributes \small{$S_j \in S$}\normalsize~across all classes is less than \small{$\mu$}\ \normalsize  i.e., $\eqDM_{S} \le \mu$.

\end{definition}
\normalsize
\noindent In the ideal case, \small{$\mu$}\normalsize=0 which signifies no discrimination w.r.t. any protected attribute and in any class.

\section{Multi-Fairness-aware Learning}
\label{sec:ourMethod}
Our goal is to learn a \eqDM-fair classifier: $f(\cdot):X \rightarrow Y$ that achieves \emph{equal low} error rates for all the groups ($g_j/\overline{g_j},~j=1,\cdots,k$) in both the classes ($+/-$). 
To this end, we first formulate clear objectives (Sec.~\ref{sec:ourFormulation}) and  then,  propose a sequential learner approach to find $f(\cdot)$ (Sec.~\ref{sec:ourAlgorithm}).

\subsection{Multi-discrimination-free Learning under Class-imbalance}
\label{sec:ourFormulation}
We define three objectives for the \eqDM-fair classifier $f(\cdot)$: 
low overall error ($O_1$),
similar (low) error rates across all classes ($O_2$), and mitigation of discriminatory outcomes for all protected attributes ($O_3$).

Objective $O_1$ targets overall error and is defined as minimizing the classification loss (0-1 loss):
\small
\begin{equation}\label{eq: err_loss}
    \begin{aligned}
     \phantom{123123}O_1:~ L(f)=\frac{1}{n}\sum_{(x_i,y_i)\in D}{|y_i-f(x_i)|}\phantom{123123123}
    \end{aligned}
\end{equation}
\normalsize
where $f(x_i)$ is the predicted and $y_i$ is the true class of $x_i$.

Objective $O_2$ explicitly targets class-imbalance by ensuring balanced performance across both classes.
Motivated by~\cite{garcia2008new_balancedloss}, we define a balanced loss function to minimize the performance differences between the two classes:
\small
\begin{equation}\label{eq:bal_loss}
\begin{aligned}
O_2:~ B(f)=|\phantom{12} \frac{1}{|D_+|}\sum_{(x_i,y_i)\in D_+}{|y_i-f(x_i)|}-\frac{1}{|D_-|}\sum_{(x_i,y_i)\in D_-}{|y_i-f(x_i)|}\phantom{12}|
\end{aligned}
\end{equation} 
\normalsize
where $D_Y \subset D$, $Y \in \{+,-\}$ denotes the set of instances belonging to class $Y$.

$O_3$ is the \mf~objective aiming to mitigate discrimination due to multiple protected attributes $S_j \in S$ and across both classes. We call it $\eqDM_S$ loss, as on optimization it aims to mitigate $\eqDM_S$ (c.f. Def.~\ref{def:eqDM}): 
\small
\begin{equation}
    \begin{aligned}\label{eq: phi_m}
    O_3:~ \Phi(f)=   \max_{S_j\in S}\big(\max_{Y\in\{+,-\}} (|\frac{1}{|g_{jY}|}\sum_{(x_i,y_i)\in g_{jY}}|y_i-f(x_i)|
    -\frac{1}{|\overline{g_{jY}}|}\sum_{(x_i,y_i)\in \overline{g_{jY}}}|y_i-f(x_i)|~|)\big )
    \end{aligned}
\end{equation}
\normalsize


\noindent where $|g_{jY}|$ is the cardinality of group $g_j$ in class $Y \in \{+,-\}$. 

The objectives $O_3$ and $O_2$  ensure similar performance across all the protected/non-protected groups and the (+/-) classes respectively, thus minimizing the performance bias against the underrepresented protected groups in the minority (+) class, while the objective $O_1$ would help establish low error rate overall. 

\subsection{The \eqDM-fair  Boosting Post Pareto (\Multifair)~Algorithm}
\label{sec:ourAlgorithm}

Our goal is to develop a classifier that takes into consideration the above three objectives, eventually solving the problem of \mf~under class-imbalance. Boosting-based \cite{shapire1999brief} approaches have been promising in tackling class-imbalance~\cite{chawla2003smoteboost,sun2007cost} and discrimination~\cite{hickey2020shapECML20fairness,IosNto:CIKM:19}. 
However, they have also been criticised for being vulnerable in the presence of noise or outliers. As outliers are more likely to be missclassified, boosting may overshoot over the iterations the weights of those instances \cite{li2018boostingOutlier}.
Thus, the ensemble obtained at the end of a predefined number of boosting rounds may produce inferior outcomes than an ensemble produced in an earlier round.

Inspired by the literature, we propose a boosting-based learner that \emph{in-training} modifies the distribution weights to incorporate our objective goals.
The new weighting puts more attention to the instances from the protected groups in minority class (as they are frequently missclassified) and might therefore, aggravate the weight overshooting problem.
To overcome this drawback, we deploy a \emph{post-training} step to select the best solution (partial ensemble).  

\subsubsection{In-training: \eqDM-boosted weight distribution update.}

\noindent Let $T$ be the number of boosting rounds. In each round $t$ $(1\leq t\leq T)$ we train a weak learner (a decision stump) based on the current instance weight distribution $D_t$. In the first round, all instances receive the same weight: $D_1(x_i)=\frac{1}{n}$. In a later  round $0 <t+1 \leq T-1$, the weight distribution is updated as follows:
\begin{equation}
\label{eq:weight_update}
 D_{t+1}(x_i)=\frac{D_{t}(x_i) \exp(-\alpha_{t} sign(y_ih_{t}(x_i))) {fc}_{t}(x_i)}{Z_{t}}   
\end{equation}

where as in AdaBoost
$\alpha_t=\frac{1}{2}\ln\frac{1-\sum_n D_t(x_i)}{\sum_n D_t(x_i)}$ is the weight of the weak learner $h_t$, $sign(y_ih_t(x_i))$ returns $-1$ if $h_t(x_i)\ne y_i$ and $1$ otherwise, and 
$Z_t$ is the normalization factor which ensures that $D_{t+1}$ is a probability distribution. The term ${fc}_{t}(x_i)$ is our modification, which corresponds to the \mf~cost (MMM-cost) for a misclassified instance $x_i$ 
defined as: 
\begin{equation}
\label{s_phi}
\phantom{123}{fc}_{t}(x_i)=
\begin{cases}
        \max_{1 \leq j \leq k}({cdc}_{ij}^{(t)}),~\textit{if}~h_t(x_i)\ne y_i\\
        1,\textit{otherwise}\phantom{123123}
\end{cases}
\end{equation}
where ${cdc}_{ij}^{(t)}$ is the discrimination weight of instance $x_i$ at round $t$  
concerning protected attribute $S_j$, which depends on the group membership of $x_i$ w.r.t $S_j$. It is defined as:
\small
\begin{equation}\label{fcm}
{cdc}_{ij}^{(t)}=
  \begin{cases}     
  \begin{aligned}
   1+|\delta FNR_j^{1:t}|,~\textit{if}~(\delta FNR_j^{1:t} \ge 0~\wedge x_{i}\in g_{j+})
   \vee (\delta FNR_j^{1:t} \le 0~\wedge x_{i}\in \overline{g_{j+}}) ;  
  \end{aligned}
  \\
    \begin{aligned}
        1+|\delta FPR_j^{1:t}|,~\textit{if}~(\delta FPR_j^{1:t} \ge 0~\wedge x_{i}\in g_{j-})
        \vee (\delta FPR_j^{1:t} \le 0~\wedge x_{i}\in \overline{g_{j-}});
    \end{aligned}  
    \end{cases}
\end{equation}\normalsize
where \small$\delta FNR_j^{1:t}$\normalsize~and \small$\delta FPR_j^{1:t}$\normalsize~are the cumulative discrimination of the partial ensemble \small$H_t(x_i)=\sum_{l=1}^t \alpha_l h_l(x_i)$\normalsize~for $S_j$ as in~\cite{IosNto:CIKM:19}. 

In each boosting round $t$ we evaluate the partial ensemble  $H_t$ 
and collect the solution vector $\vec{f_t}=[o_1, o_2, o_3]_t$, 
where $o_i=O_i(t)$ is a solution point of $H_t$ for the respective objective $O_i$. In total, $T$ solution vectors are collected. 
The sequential training stops when the maximum number of iterations $T$ is reached. 

\subsubsection{{Post-training: Selecting Pareto Optimal Solution.}} 
\noindent Our goal is to find the optimal round $t^*\leq T$ to output the partial-ensemble with the best $(O_1,O_2,O_3)$ objectives trade-off:
$$H_{t^*}=\sum_{l=1}^{t^*}\alpha_l h_l$$

This is achieved in two steps: First, out of all $T$ solutions we select the set of non-dominating optimal solutions. 
 Next, we find the best trade-off solution among the shortlisted ones to get the corresponding optimal~$t^*$.

\paragraph{\underline{\textbf{1. Pareto front computation:}}}
\noindent Among all solution vectors $\vec{f_t},~t=1,\cdots,T$ collected over the boosting rounds, we find the Pareto Front ($\mathbb{PF}$), i.e., the non-dominated set of Pareto optimal solutions. A solution $\vec{f_{t'}}$ is said to be dominated by a solution $\vec{f_t}$ if \small$1)~O_i(t)\le O_i(t')~\forall i\in\{1,2,3\}$, and $2)~ \exists i\in\{1,2,3\}~O_i (t) < O_i(t')$.\normalsize 

\paragraph{\underline{\textbf{2. Pseudo-weight calculation and choice of best solution}}}:
\noindent To choose the best solution we use the pseudo-weight algorithm~\cite{deb2001multi__pseudoweights} that calculates the relative distance of each solution from the worst (maximum value) solution 
for each objective. 
The pseudo-weight $w_{ti}$ for $o_i \in \vec{f_t}$ is given by:
\begin{equation}\label{eq:pseudo_w}
    \begin{aligned}
        w_{ti}=\frac{(o_i^{\max}-o_{ti})/(o_i^{\max} - o_i^{\min}) }{ \sum_{i=1}^{3} (o_i^{\max}-o_{ti})/(o_i^{\max} - o_i^{\min}) }
    \end{aligned}
\end{equation}
where $o_i^{\max}$ ($o_i^{\min}$) is the maximum or worst (minimum or best) objective value achieved in any of the rounds. 
This way, for each solution $\vec{f_t}=[o_1,o_2,o_3]_t$ we compute the corresponding pseudo-weight vector $\vec{w_t}=[w_{t1},w_{t2},w_{t3}]$. 
Next,
we select the solution with the least relative weighted sum as the best trade-off solution w.r.t all the objectives:
\small
\begin{equation}\label{eq: relsum}
   \vec{f_t^*}= argmin_{t} \{(1-\vec{w_t})\cdot \vec{f_t}\}= argmin_{t}\{\sum_{i=1}^3 (1-w_{ti})o_{ti}\}
\end{equation}
\normalsize
where $(1-\vec{w_t})$ is the required transformation as the pseudo-weights vector by its nature 
assigns bigger weight $w_{ti}$ to a smaller objective solution value $o_{ti}$.

\section{Experiments}
\label{sec:experiments}
We evaluate \Multifair
~performance against state-of-the-art approaches 
(Sec.~\ref{sec:exp_performance}). 
To show the utility of our \eqDM-cost (Eq.~\ref{s_phi}) in tackling balanced error ($O_2$), we plot balanced loss $B(f)$ with varying \eqDM~tolerance thresholds $\mu$ (Sec.~\ref{sec:analysis}). Further, we show the changes in the dataset distribution over training and the effectiveness of our approach in promoting underrepresented protected groups (Sec.~\ref{sec:analysis}).  
We plot the $O_1,O_2,O_3$ losses over the rounds to justify the need for post-training selection.
At last, in Sec.~\ref{sec:flex} we show the flexibility of  \Multifair~to intake user preferences for post-training selection.

\subsection{Experimental settings}
\label{sec:exp_settings}

\noindent\paragraph{\textbf{Baselines:}} We compare against four state-of-the-art fairness-aware methods:

\noindent\textit{FairCons}~\cite{zafar2019fairness}: Tackles \mf~fairness-related convex-concave constraints, 

\noindent\textit{FairLearn}~\cite{agarwal2018reductions}: imposes a set of linear fairness constraints on an exponentiated-gradient reduction method to tackle \mf, 

\noindent\textit{MiniMax}~\cite{martinezminimax}: tackles \mf{} as a mini-max game while searching for a Pareto efficient solution of a multi-objective problem,

\noindent\textit{W-ERM}~\cite{yang2020fairness_overlap}: applies a Bayes-optimal group-fair classifier to consider algorithmic fairness across multiple overlapping groups simultaneously to tackle the \mf~trade-off,

\noindent\textit{MI-Fair}~\cite{kang2021multifair_MI}: minimizes mutual information between the learning error and the vectorized multiple protected attributes to tackle \mf,  and 
\noindent\textit{AdaFair}~\cite{IosNto:CIKM:19}: uses \sd~based boosting algorithm along with summed accuracy and class-imbalance loss to tackle \sd~and class-imbalance. 

In order to understand the effect of the post-training part as well as the effect of the  $\mathbb{PF}$ selection in the post-training part, we also include in the experiments two variations of \Multifair: 
\begin{itemize}
    \item \textit{MFB} that completely discards the post-training part and \item \textit{MFBP} that uses post-training but does not use the Pareto Front $\mathbb{PF}$ set for the final selection but rather selects from all solutions.
\end{itemize}

\noindent\paragraph{\textbf{Datasets:}}\

We report on three imbalanced real-world datasets (c.f., Table \ref{tab:data}). Additionally, we also report on Compas~\cite{larson2016we} (CIR: $1:1.2$) to show the usability of our method also for class-balanced scenarios. 
The protected attributes and protected groups studied in the experiments are \emph{Sex} ($g_j$=``female"), \textit{Race} ($g_j$=``non-white"), \textit{Marital status/Mari} ($g_j$=``married"), \textit{Age} ($g_j$=``$\leq25 \And \geq 60$").
         

\noindent\paragraph{\textbf{Evaluation Measures:}}\

For $O_1$, we report on accuracy ($Acc$), for $O_2$ on geometric mean ($G.M$) and for $O_3$ we report on the proposed \emph{\eqDM}-fairness, as well as on the $DM$ for each protected attribute. Additionally, we report on the accuracy of 
the worst 
performing protected group in the minority (+) class (${Wg}_{+}$).

\noindent\paragraph{\textbf{Experimental setup:}}\

We set the number of weak learners to $T=500$. We follow the same
evaluation setup as in \cite{IosNto:CIKM:19,zafar2019fairness} by splitting each dataset randomly
into train (50\%) and test (50\%) and report on the average of 10
random splits. 

\subsection{Evaluation results}
\label{sec:exp_performance}
\begin{figure}
    \centering
    \includegraphics[width=0.9\columnwidth]{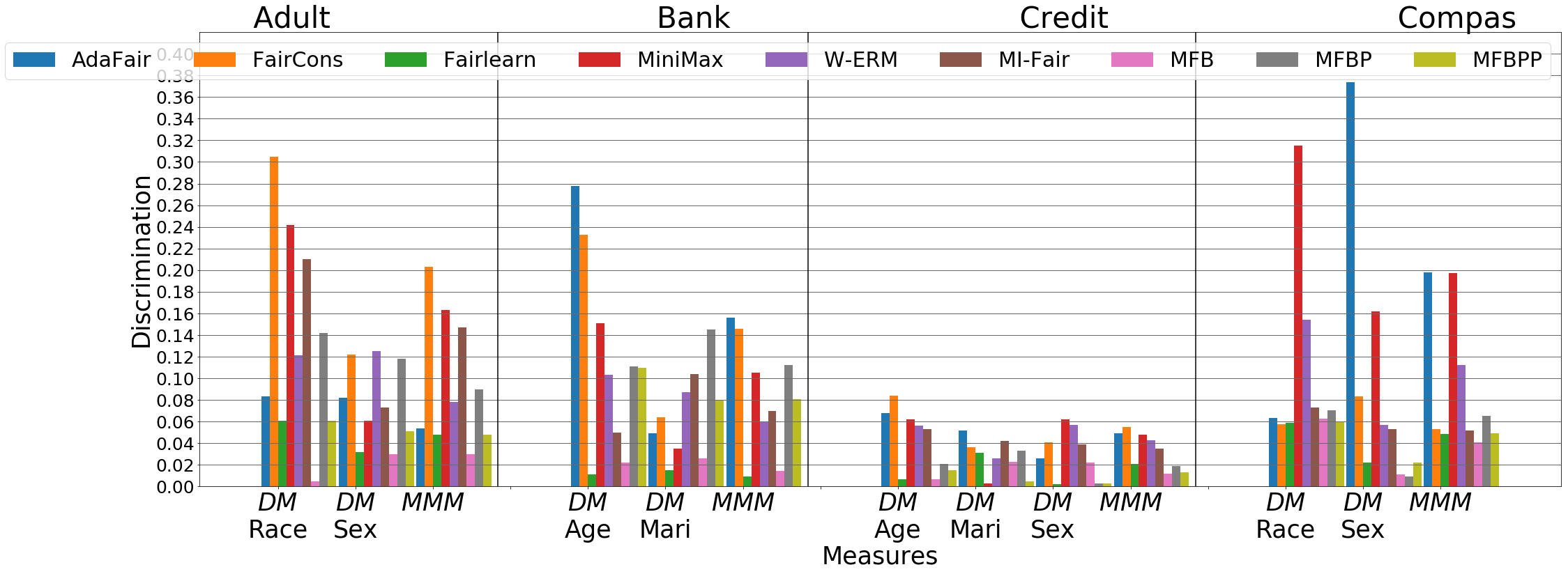}
    \caption{ Discrimination performance: For each dataset, the overal \eqDM~score and the DM scores for each protected attribute are shown (lower values are better).}
    \label{fig:fairness}
    \vspace{-4mm}
\end{figure}

The discriminatory and predictive performance evaluation of the different approaches is shown in Fig~\ref{fig:fairness} and  Table~\ref{tab:predict}, respectively.

    
\noindent 

\begin{table}
\vspace{-4mm}
    \centering
    \caption{Predictive performance evaluation. $W_{g+}$ is the accuracy of the worst performing protected group in the minority (+) class }    \label{tab:predict}
    \begin{tabular}{|c|c|c|c|c|}
    \hline
         &\textbf{Adult} &\textbf{Bank} &\textbf{Credit} &\textbf{Compas}   \\
         &\small ${Acc}$~~~${Wg}_{+}$~~~$G.M$ &\small ${Acc}$~~~${Wg}_{+}$~~~$G.M$ &\small ${Acc}$~~~${Wg}_{+}$~~~$G.M$ &\small ${Acc}$~~~${Wg}_{+}$~~~$G.M$ \normalsize \\
         \hline
         AdaFair &0.84~~~0.63~~~0.76  & 0.88~~~0.62  ~~~0.76 &  
         0.81~~~ 0.33 ~~~0.57 &  0.65~~~ 0.50 ~~~0.64\\
         FairCons &0.85~~~0.43~~~0.75  & \textbf{0.91~}~~0.33  ~~~0.59 &  
         0.81~~~ 0.28 ~~~0.55& 0.67~~~ 0.53 ~~~0.66 \\
         FairLearn &0.83~~~0.54~~~0.73  & 0.88~~~0.21  ~~~0.46 &  
         0.79~~~ 0.21 ~~~0.45& 0.65~~~ 0.55 ~~~0.64 \\
         MiniMax &\textbf{0.86}~~~0.49~~~0.76  & 0.90~~~0.45  ~~~0.66 &  
         \textbf{0.82}~~~ 0.37 ~~~0.60&  \textbf{0.68}~~~ 0.49~~~0.67\\
         W-ERM & 0.85~~~0.52~~~0.75 & 0.90~~~0.29  ~~~0.56 &  
         0.81~~~ 0.22 ~~~0.47 & 0.66~~~0.47~~~0.64  \\
         MI-Fair & 0.84~~~0.65~~~0.72 & 0.89~~~0.69  ~~~0.82 &  
         0.80~~~ 0.59 ~~~0.68 & 0.68~~~0.60~~~\textbf{0.68}  \\
         \hline
         MFB &0.69~~~0.64~~~0.74  & 0.36~~~0.28   ~~~0.41&  
         0.71~~~ 0.68 ~~~0.70 &0.64~~~ 0.63 ~~~0.63  \\
         MFBP &0.85~~~0.77~~~\textbf{0.84}  & 0.85~~~\textbf{0.75}   ~~~\textbf{0.85} &  
         0.71~~~ 0.64 ~~~0.69 &0.67~~~ 0.60 ~~~0.65  \\
         \Multifair & 0.81~~~\textbf{0.79}~~~0.81 & 0.81~~~0.72  ~~~0.80 &  
         0.74~~~ \textbf{0.65} ~~~\textbf{0.70} & 0.66~~~ \textbf{0.63} ~~~0.66 \\
         \hline
    \end{tabular}
    
\end{table}

\noindent{\bf \emph{Multi-discrimination}:} From Fig.~\ref{fig:fairness} we notice that our MFB outperforms all the approaches in all the dataset. \Multifair~comes second outperforming the baseline competitors in mitigating \mf~(i.e., objective $O_3$) by producing the lower $\eqDM$ discrimination values in three datasets (Adult: \textbf{0.05}, Compas: \textbf{0.04}, Credit: \textbf{0.01}), while falling behind FairLearn in one dataset (Bank: 0.08). However, in Table~\ref{tab:predict} we notice that MFB severely underperforms in all the predictive evaluation measures, thus failing to provide a good trade-off between $O_1,O_2,$ and $O_3$. 
The closest competitor to us w.r.t. fairness is FairLearn, which however achieves low discrimination by consistently ignoring the minority class (A closer look to  Table~\ref{tab:predict}, shows that FairLearn achieves the lowest $G.M$ for all four datasets). 
Approaches like FairCons, MiniMax, and W-ERM result in different levels of discrimination for the different protected attributes and overall high \eqDM~values. MI-Fair have mixed outcome with high discrimination in Adult, but performed at par with \Multifair~in Bank and Compas data. AdaFair
trained on one protected attribute (for Adult: sex, for Bank: marital status, for Credit: sex, for Compas: race) does not mitigate discrimination for other protected attributes and consequently also results in high \eqDM~values esp. for Bank and Compas. In case of Adult and Credit datasets, AdaFair, albeit trained for \sd{} it seems to tackle \mf; the reason is the strong correlation between the protected attributes as revealed by chi-square test with $\rho$-$value\approx 0$. 

\noindent{\bf Underrepresented protected groups ($g_{j+}$):} In Table~\ref{tab:predict} we notice that \Multifair{} and {MFBP} both outperform the other approaches on ${Wg}_{+}$ by far ($[5\% - 21\%\uparrow])$. Thus, our proposed methods overcome the issue of bias due to the imbalanced distribution of protected groups (c.f Table~\ref{tab:data}), ensuring high predictive accuracy for any $g_{j+}$. Note that all the other approaches that even after mitigating \mf{} fail on this task. MI-Fair emerges as the best among the baseline competitor in all the four datasets behind our proposed methods \Multifair, MFBP,and MFB in Adult, Credit, and Compas datasets, while outperforming only MFB in Bank data. 

\noindent{\bf Balanced performance:}  Table~\ref{tab:predict} shows that our MFPB and MFBP outperform the baselines in $G.M$ in the range $[4\% - 11\%] \uparrow$ for the imbalanced Adult, Bank, and Credit datasets, while being marginally behind MI-Fair, and Minimax in the balanced  Compas dataset. We can easily notice that our $Acc$ and $G.M$ values are close to each other for all 
the datasets with $Acc/G.m\approx 1$. This indicates we achieve $B(f)\approx 0$ ($O_2$), in all the datasets. Our closest competitors here are MI-Fair, AdaFair and MiniMax. AdaFair explicitly targets class-imbalance for mono-discrimination. MiniMax, and MI-Fair indirectly tackles the problem as they aims at minimizing error for all groups.  For other baselines, $Acc/G.m>> 1$, indicating substantial performance differences between the classes.

\noindent{\bf Overall accuracy:} 
\Multifair~is marginally compromised on the overall $Acc$ in Adult, Credit, and Bank datasets ($[12\% - 5\%] \downarrow$). MiniMax emerged as the winner here, accomplishing the best accuracy in Adult, Credit, and Compas dataset. This is the trade-off we pay to ensure  
nearly equal 
performance for all (protected/non-protected) groups across all the classes. 
\begin{figure}
    \centering
    \includegraphics[width=0.99\textwidth]{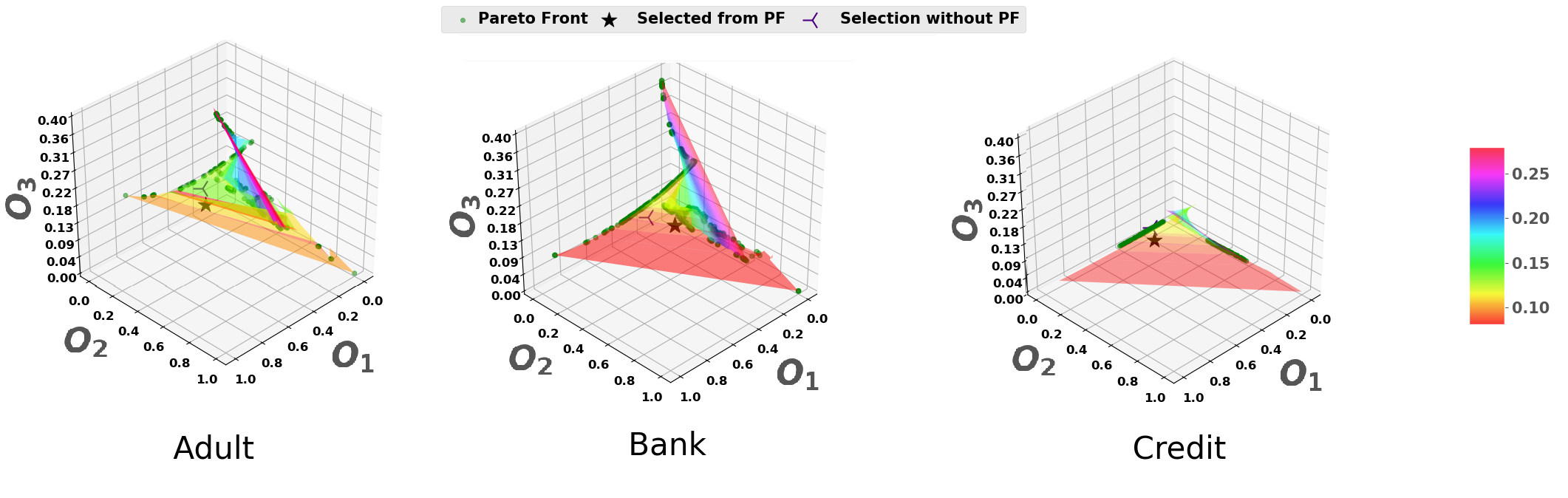}
    \caption{Visualization of the Pareto front and the selected trade-off solution in the complete solution surface.}
    \label{fig:pareto_explain}
\end{figure}

\noindent {\bf Summary:} \Multifair{} provides the best holistic outcome in overall trade-off, outperforms the baselines in mitigating \mf, produces the best predictive performance on underrepresented protected groups ($\forall_j g_{j+}$) in minority class, and equal performance across all classes, while maintaining comparable high accuracy against the baselines. 
MFB produces the most-fair outcomes but suffers in predictive performance. It overshoots the weight and increases overall error to gain fairness. MFBP solves the overshooting problem but gets outperformed by \Multifair{} in the fairness task. 
In Fig~\ref{fig:pareto_explain} we see that the solution surface after the training-MFB phase of our algorithm 
is very wide spread in the $O_1,O_2,O_3$ objective space. 
By computing $\mathbb{PF}$ 
as in \Multifair, we 
narrow down the search space.
Using the pseudo-weights, we pick a solution each time close to the origin in the objective space (which is desired). Hence, we are always able to deliver a good trade-off solution without any hyper-parameter tuning. 
FairLearn also tackled the \mf~problem consistently well, but by under-performing in the minority class. MiniMax lacks in \mf{} convergence but produces the most overall accurate ($O_1$) predictions.  FairCons has difficulty in finding the optimal parameters leading to its poor \mf{} performance. W-ERM apart from Bank dataset (the most imbalanced), always delivers comparable trade-offs. However, the method is very slow. MI-Fair can be argued as the closest competitor in overall trade-off delivering balanced and accurate performance with low discrimination in three out of the four datasets under study. 


\subsection{Internal analysis}\label{sec:analysis}
This section aims to analyse \Multifair's ability to produce state-of-the-art balanced performance while dealing with \mf. In particular, we try to find  answers for three significant points: i) How \Multifair~ensures high accuracy for the underrepresented protected groups in the imbalanced minority (+) class? ii) How the overshooting problem affects and, is the post-processing step really required?  iii) Does the \mf~cost (Eq.~\ref{s_phi}) also tackle the balanced loss and, what happens if we relax cost by varying the \eqDM~threshold $\mu$ (Def.~\ref{def:MMM-fairlearner})? 
Here we focus the study using only the imbalanced data (Table~\ref{tab:data}).

\begin{figure}
    \centering
    \includegraphics[width=0.9\columnwidth]{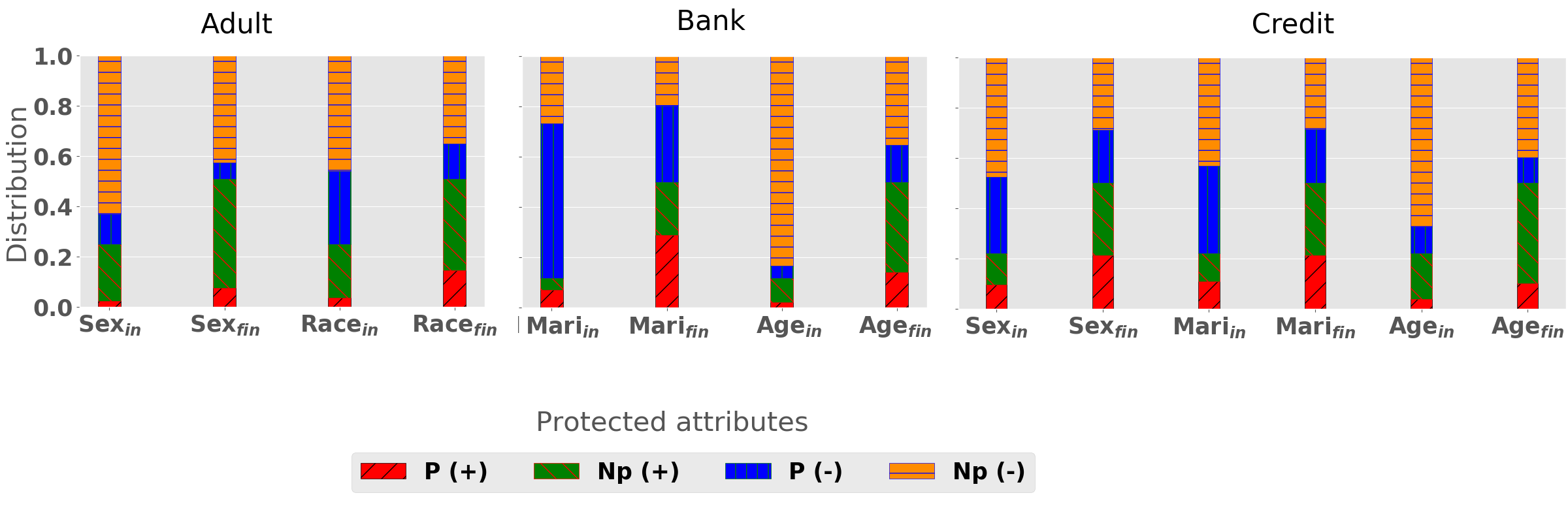}
    \caption{ Changes in instance weight distribution. For each dataset and protected attribute $S_j$, we depict the initial distribution $S_{in}$ and the final one $S_{fin}$.}
    \label{fig:weight_dist}
    \vspace{-4mm}
\end{figure}
\noindent\textbf{Answer to point (i)}: We analyse the changes in weight distribution of the various groups from its initial (\textit{ini}) distribution (actual data representation), to boosted weight till the finally selected partial ensemble point (\textit{fin}) in Fig~\ref{fig:weight_dist}. For any protected attribute $S_j$, $P$ and $Np$ refer to the respective protected and non-protected groups. Thus, $P(+)$ translates as the protected group in the minority (+) class  ($g_{j+}$). 
We notice that \textit{ini} weights of each $P(+)$ in every dataset is largely underrepresented. But, in the \textit{fin} weights each $P(+)$ group is boosted significantly. \Multifair~increases the weight of the underrepresented groups, thus changing the decision boundary to produce highly accurate and unbiased results for all the groups even in case of high imbalance. 

\begin{figure}
    \centering
    \includegraphics[width=0.9\columnwidth]{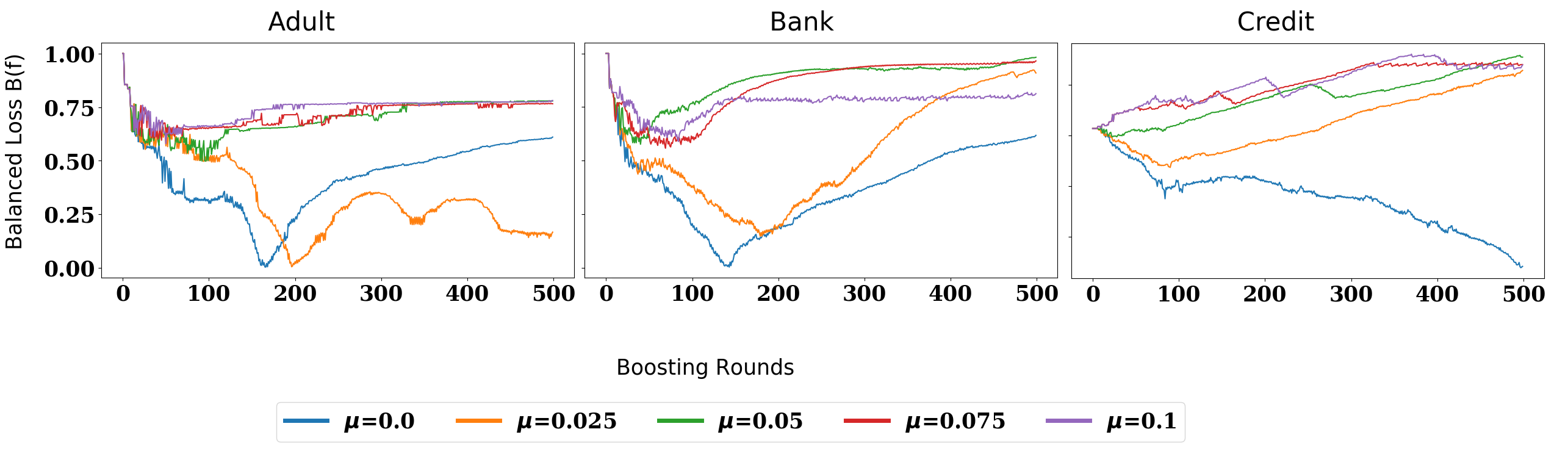}
    \caption{ B(f) loss over boosting rounds with varying \eqDM~thresholds $\mu$.}
    \label{fig:o2_over_rounds}
    \vspace{-4mm}
\end{figure}

\noindent\textbf{Answer to points (ii):}
We have already shown  
the effectiveness of our \eqDM{} cost (Eq.~\ref{s_phi}) in mitigating \mf. Now to understand its effect on balanced loss 
we monitor $B(f)$ over the boosting rounds (Fig~\ref{fig:o2_over_rounds}) for different MMM-tolerance thresholds $\mu$.
We see that when $\mu=0$, the shape of the $B(f)$ loss curve is parabolic for Adult and Bank datasets. In Credit data, the loss continues to descent till the final round. The parabolic curve supports our intuition of the possibility of overshooting the weights due to the possible repeated boosting of noisy instances, whereas a consistently descending loss curve for Credit data shows the uncertainty involved in estimating the optimal size of the ensemble. 
These results justify the necessity of the  post-training selection part. 

\noindent\textbf{Answer to points (iii):} In Fig~\ref{fig:o2_over_rounds} we also show the effect of different \eqDM~threshold $\mu$ values on $B(f)$.
By increasing $\mu$ we relax the \eqDM~boost i.e we have ${fc}_t(x_i)=1$ in Eq.~\ref{eq:weight_update} when discrimination (Eq.~\ref{s_phi}) is $\leq (1+\mu)$. We observe the immediate effect on the $B(f)$ loss. In each of the datasets, the effectiveness of \Multifair~to tackle class imbalance decreases as the $B(f)$ loss increases while we increase the threshold $\mu$. Thus, showcases the ability of our \eqDM~cost in tackling the $O_2$ along with our \mf~objective $O_3$.

\subsection{Flexibility of \Multifair}\label{sec:flex}
Thus far, we use the pseudo-weight method (Eq.~\ref{eq: relsum}) to select the best solution among the ($\mathbb{PF}$) solutions.
If information on user preferences exists, in the form of a user-preference vector $\small \vec{u}=[u_1,u_2,u_3]\normalsize$: $\small u_1 +u_2+u_3 = 1\normalsize$, it can be used to select the best solution according to user needs. In this case, we choose the solution 
$\vec{f_{t^*}}$ whose corresponding pseudo-weight $\vec{w_{t^*}}$ is closest according to L1 distance, to the preference vector $\vec{u}$.  
To evaluate the effect of such an approach, we mimic four different users and provide their preference vector $\Vec{u}$ as an additional input to \Multifair. In particular, we assume the following users: 
i) $\Vec{u}=[0.33,0.33,0.33]$ indicating \emph{equal preference} to all $O_i$,
ii) $\Vec{u}=[0,0,1]$, 
iii) $\Vec{u}=[0,1,0]$, 
iv) $\Vec{u}=[1,0,0]$, indicating preference only for $O_i$ if $u_i=1$.

\begin{figure}
\vspace{-4mm}
    \centering
    \includegraphics[width=0.9\columnwidth]{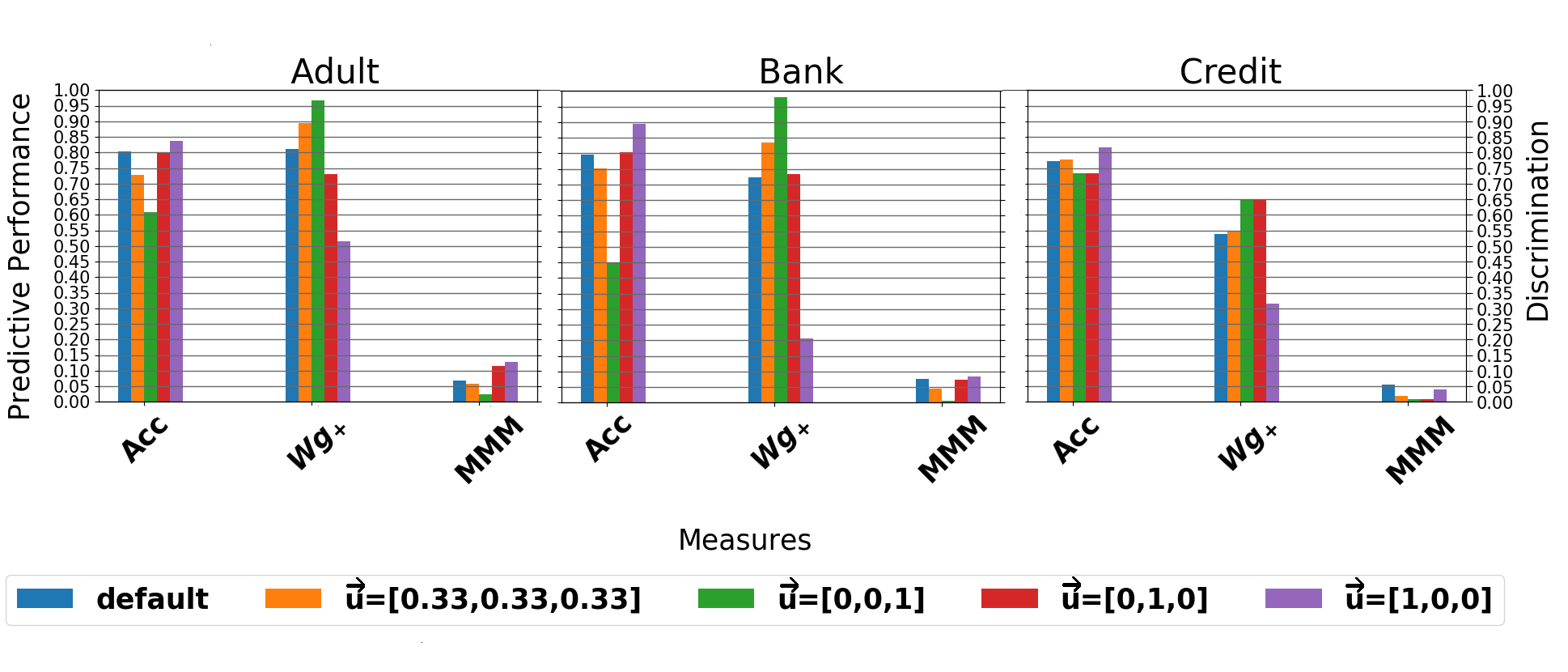}
    \caption{Performance evaluation for different user preference vectors $\Vec{u}$}
    \label{fig:preference}
    \vspace{-4mm}
\end{figure}

As expected, the output changes noticeably with changes with $\Vec{u}$. For all datasets the most accurate classifier (for $\Vec{u}=[1,0,0]$) delivers Acc at par if not better than the state of the art, whereas the fairest (for $\Vec{u}=[1,0,0]$) produces state of the art fair predictions. With preference $\Vec{u}=[0.33,0.33,0.33]$ the classifier consistently produces good trade-off solutions, however, the default version of \Multifair~(without $\Vec{u}$) produces better trade-offs.

\section{Conclusions and Outlook}
\label{sec:conclusions}
In this work we claimed that \mf{} under class-imbalance is an important multi-faceted problem of finding low overall error, while minimizing performance differences across the classes and groups.
Existing multi-discrimination approaches consider only error-discrimination trade-off, and
ignore class-imbalance. This way, they
achieve \mf{} by under-performing in the minority (+) class, especially for the underrepresented protected groups. To this end, we propose the \multimax{} fairness measure (\eqDM) and a \eqDM-fair boosting post Pareto classifier (\Multifair) to ensure \eqDM-fairness. 
Our experiments show 
the superiority of our method in mitigating \mf, producing best balanced performance across groups and classes  along with best accuracy for underrepresented protected groups in the minority (+) class, without a significant compromise on overall accuracy.
Further, our method is flexible to user needs 
as it can select the best 
solution trade-off
according to user preferences. In future, we want study the \mf{} under class-imbalance problems in the more challenging multi-class and multi-label set-up, where the complexity is much harder.

\section*{Acknowledgements}
The work of the first author is supported by the Volkswagen Foundation under the call ``Artificial Intelligence and the Society of the Future" (the BIAS project). We are sincerely thankful to the invaluable suggestion of Prof. Niloy Ganguly from L3S Research Center, in shaping up the paper to its current form. 
\bibliographystyle{splncs04}
\bibliography{biblio}
\end{document}